\documentclass{article}

\usepackage[preprint]{neurips_2025}

\usepackage{booktabs}
\usepackage{graphicx}
\usepackage{adjustbox}
\usepackage[utf8]{inputenc} % allow utf-8 input
\usepackage[T1]{fontenc}    % use 8-bit T1 fonts
\usepackage{hyperref}       % hyperlinks
\usepackage{url}            % simple URL typesetting
\usepackage{booktabs}       % professional-quality tables
\usepackage{amsfonts}       % blackboard math symbols
\usepackage{amsfonts}       % blackboard math symbols
\usepackage{xspace}         % fixes \ie undefined control sequence
\usepackage{nicefrac}       % compact symbols for 1/2, etc.
\usepackage{microtype}      % microtypography
\usepackage{xcolor}         % colors
\usepackage{wrapfig}
\usepackage{enumitem}
\usepackage{amsfonts}
\usepackage{amssymb}
\usepackage{bbm}
\usepackage[many]{tcolorbox}
\usepackage{colortbl}
\usepackage{multirow}
\usepackage{arydshln}
\usepackage{natbib}
\usepackage{tabularray}
\usepackage{neurips_2025}

\usepackage{amsmath}
\usepackage{amssymb}
\usepackage{mathtools}
\usepackage{amsthm}
\usepackage{float}
\usepackage[capitalize,noabbrev]{cleveref}

\theoremstyle{plain}
\usepackage{booktabs}
\usepackage{adjustbox}
\usepackage{array}  % for p{} column types

\theoremstyle{definition}

\theoremstyle{remark}

 \def \VersionWithComments {}
\ifdefined 
\VersionWithComments
\usepackage{marginnote}

\usepackage[textsize=tiny]{todonotes}

\def \VersionWithComments {}
\ifdefined 
\VersionWithComments
\usepackage{marginnote}

\definecolor{mygray}{gray}{.9}

\makeatletter
\DeclareRobustCommand\onedot{\futurelet\@let@token\@onedot}
\def\@onedot{\ifx\@let@token.\else.\null\fi\xspace}
\def\eg{e.g\onedot,\xspace} \def\Eg{\emph{E.g}\onedot,\xspace}
\def\ie{i.e\onedot,\xspace} 
 
\def\etc{etc\onedot} 
 
\def\etal{et al\onedot}

\title{Towards Universal \& Efficient Model Compression via Exponential Torque Pruning}

\author{%
  \parbox{\textwidth}{\centering Sarthak K.~Modi$^{1}$,\quad Lim Zi Pong$^{2}$, \quad Shourya Kuchhal$^{1}$, \quad Yushi Cao$^{1}$ \quad Yupeng Cheng$^{1}$, \quad Yon Shin Teo$^{2}$, \quad Shang-Wei Lin$^{3}$, \quad Zhiming Li$^{1,}$\thanks{Corresponding Author}} \\
  $^{1}$ Nanyang Technological University\\
  $^{2}$ Continental Automotive Singapore\\
  $^{3}$ Singapore Institute of Technology\\
  \texttt{\{sarthakk001,Zhiming001\}@e.ntu.edu.sg} \\
}

\begin{document}

\maketitle

\begin{abstract}
The rapid growth in complexity and size of modern deep neural networks (DNNs) has increased challenges related to computational costs and memory usage, spurring a growing interest in efficient model compression techniques. 
Previous state-of-the-art approach proposes using a Torque-inspired regularization which forces the weights of neural modules around a selected pivot point. Whereas, we observe that the pruning effect of this approach is far from perfect, as the post-trained network is still dense and also suffers from high accuracy drop. In this work, we attribute such ineffectiveness to the default linear force application scheme, which imposes inappropriate force on neural module of different distances. To efficiently prune the redundant and distant modules while retaining those that are close and necessary for effective inference, in this work, we propose \underline{E}xponential \underline{T}orque \underline{P}runing (ETP), which adopts an exponential force application scheme for regularization. Experimental results on a broad range of domains demonstrate that, though being extremely simple, ETP manages to achieve significantly higher compression rate than the previous state-of-the-art pruning strategies with negligible accuracy drop.

% In this work, we first analyze the complexity of the pruning process and provide proof of it being a NP-Hard problem. Using this key insight, we propose Self-Organizing Regularization for Trailing Sparsity (SORTS), a novel loss-function-based regularization framework that promotes structured sparsity in neural networks. SORTS employs a two-stage pruning approach, minimizing the overhead of resource allocation for pruning while enabling the model itself to autonomously determine the components to prune. Extensive evaluations on several different architectures such as ResNet, VGG, MobileNet, AlexNet, DarkNet for Images, GAT for graphs, BERT-based models for language and Informers for timeseries tasks demonstrate that SORTS achieves remarkable compression rates exceeding 20×, with negligible impact on model accuracy, highlighting its potential as a practical solution for resource-constrained deep learning in variety of applications.
\end{abstract}

\section{Introduction}
\label{Intro}
% Scaling up model sizes has become a dominant trend in deep learning leading to better performance and higher sample efficiency, particularly in areas such as Generative Artificial Intelligence~\cite{scalingGAI}, Language Modeling\cite{scalingLLM}, and Computer Vision~\cite{scalingCV}. 

Deep neural networks (DNNs) have revolutionized countless domains by setting state-of-the-art baselines that significantly surpass previous approaches. However, nowadays DNNs are pretty large in size and require substantial floating point operations per second (FLOPS) for inference, which limits their applications in resource-constrained scenarios (\eg~edge devices~\cite{qin2018compress,efficientml1,efficientml2})
To achieve more efficient while also effective inference, many model compression techniques have been proposed. \Eg~low rank approximation, which aims to leverage a lower-rank matrix to capture the essential structure of the original model's weight matrix while reducing complexity~\cite{hu2022lora,visionLORA,VLMLORA}; unstructured pruning, which removes unimportant weights from DNNs to reduce its size while preserving performance~\cite{lecun1989optimal,liao2023can,unstructured}. Specifically, the previous state-of-the-art structured pruning method proposes a Torque-inspired regularization loss~\cite{torque}, named as \emph{Torque}. Concisely, analogous to the physical definition of ``Torque'', this approach uses a regularization that functions as a force that consolidates the weights of a neural module around a selected pivot point during training. By regularizing the model in this way, the weights of the neural modules that are far away from the pivot point would be forced to be zero, which could therefore be pruned.

\begin{figure}[!t]
  \centering
  \includegraphics[width=0.5\linewidth]{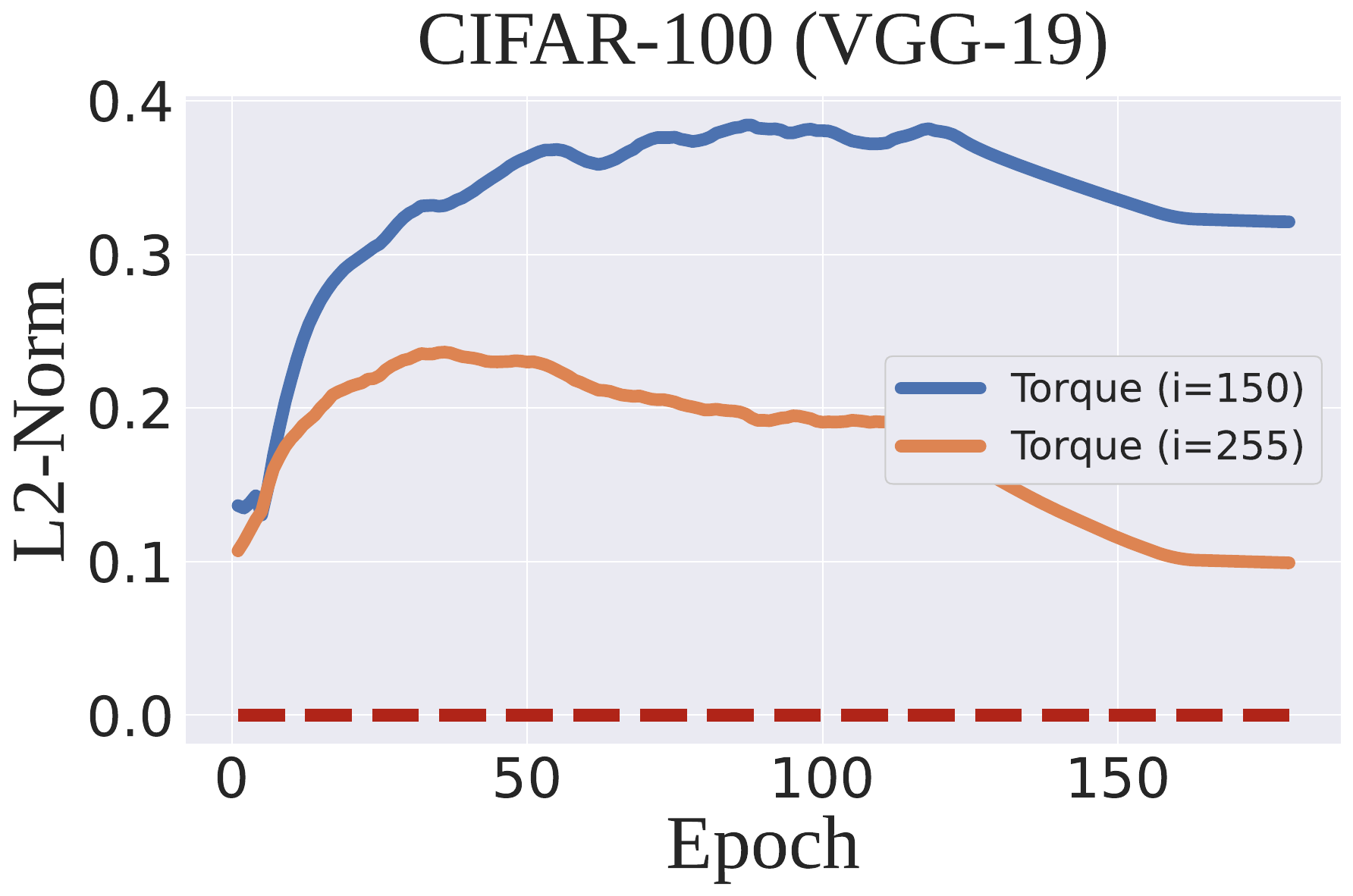}
  \caption{L2-norm curve during training process of VGG-19 on CIFAR-100.}
  \vspace{-4mm}
  \label{fig:moti}
\end{figure}

However, we observe that the vanilla Torque structured pruning is still far from perfect. \Cref{fig:moti} shows the L2-norm curves of two neural modules during the training process. Concretely, the two neural modules are of different distances $d$ from a selected pivot module, which are

all from the same network layer\footnote{The first neural module in a layer is selected to be the pivot point, and the distance of a neural module from the pivot point is number of neural modules in between}. 
It is obvious that though the two investigated modules are far from the pivot point (\ie~$d=150$, $d=255$), their L2-norm is quite high
(\ie~$||\mathbf{w}|| \gg 0$, where $\mathbf{w}$ denotes the weight tensor), which is far from optimal. Thus, because of the low sparsity of the regularized model, it still suffers from a high accuracy drop after pruning, specifically, a 7\% absolute accuracy reduction despite achieving only a $9\times$ speed-up\footnote{\emph{speed-up} is a metric that measures the reduction in computational cost, which calculates the ratio of the total operations required in the baseline model to that in the pruned model.} on the CIFAR-100 dataset for the VGG19 model. 
% Compared to this, our method achieves 23$\times$ speed-up for only 5$\%$ absolute accuracy reduction in the exact same settings. 

% Larger models, with their billions of parameters and substantial floating point operations per second (FLOPS) requirements, consistently set new benchmarks in performance in various tasks~\cite{diffusion_models,sizeML}. However, their practical deployment in resource-constrained environments, such as edge devices, remains a significant challenge due to stringent hardware, memory, and energy limitations~\cite{qin2018compress,efficientml1,efficientml2}. This challenge underscores the pressing need for advanced model compression techniques that can deliver faster, more compact neural networks while preserving their accuracy and performance~\cite{efficientml3,deng2020model,efficientml4,efficientml5}.

In this paper, we attribute the ineffectiveness of Torque to its improper force application to modules of different distances. Concisely, Torque adopts a simple linear force application scheme, which applies inadequate force on distant neural modules while imposing unnecessarily large penalties on neural modules that are close to the pivot point, which are indispensable for effective inference. To mitigate this drawback, in this work, we propose Exponential Torque Pruning (ETP), which adopts an exponential force application scheme for regularization. By applying ETP, we could efficiently prune the redundant and distant modules by applying exponentially large forces that constrain the neural modules' weights to zero while retaining those that are close and necessary for effective inference.

Though being extremely simple and straightforward, we observe that ETP manages to achieve fascinating improvements over the previous state-of-the-art baselines on various domains. Besides, ETP is universal and is directly applicable to different model architectures. For example, on the natural language understanding tasks, ETP achieves a 42$\times$ speed-up with merely 2.4\% accuracy drop on BERT (SST-2), while the previous state-of-the-art (SoTA) pruning method only achieves a 3.8$\%$ accuracy drop; and on the image classification tasks, ETP achieves a 23$\times$ speed-up with merely 4.5\% drop in accuracy on VGG-19 (CIFAR-100), while the previous SoTA suffers from a large accuracy drop of 10.8$\%$.
% a 12.16$\times$ speedup with a 0.027 reduction in F1 score on GAT (PPI), and a 24.83$\times$ speedup with a 6.15$\%$ increase in MAE for Informer(Etth1). 
% The results show that the exponential force application scheme of ETP allows it to prune the distant modules effectively while retaining those that are close and indispensable for in.

The contributions of our paper are as follows:
 \begin{itemize}
[topsep=2pt,itemsep=2pt,partopsep=0ex,parsep=0ex,leftmargin=*]
  \item We propose a universal structured pruning strategy, called Exponential Torque Pruning (ETP)\footnote{The implementation is available at: \url{https://anonymous.4open.science/r/ETP-3EB6}}, which leverages an exponential force application scheme that imposes a larger force on distant neural modules so as to constrain their weights to zero while preserving those that are close to the pivot point that are indispensable for effective inference.
  \item Experimental results on four distinct downstream domains and various model architectures validate that ETP can surpass the previous state-of-the-art pruning techniques regarding compression rate by a large margin, while retaining negligible accuracy drop.
  \item The significant improvement, high generality, and low additional training overhead pave the way for its strong potential in compressing modern Large Language Models (LLMs).
\end{itemize}

\section{Preliminary}
In this section, we introduce the fundamentals of structured pruning and a state-of-the-art structured pruning technique: torque-based structured pruning, which is the predecessor of our proposed ETP method.
\subsection{Structured Pruning}
Structured pruning is a vital technique in deep neural network compression~\cite{liu2017learning,GReg,fang2023depgraph,torque,structuredpruning_survey,centripetal_structured,efficientml4}, where entire components such as filters, neurons, or layers are removed instead of individual weights. This leads to more efficient models that are computationally and memory efficient, making them ideal for deployment on resource-constrained devices. Given a network \( \mathcal{N}(\mathbf{x}; \theta) \), where $\mathbf{x}$ denotes the input data, $\theta$ is the model's parameters, structured pruning aims to find a reduced set of parameters \( \theta^* \subset \theta \) such that:
$\mathcal{N}(\mathbf{x}; \theta^*) \approx \mathcal{N}(\mathbf{x}; \theta)$, while minimizing the network size. The pruning process is typically guided by a combined loss function:
\begin{equation}
    \mathcal{L}_{\text{total}}(\theta^*) = \mathcal{L}_{\text{task}}(\mathcal{N}(\mathbf{x}; \theta^*)) + \lambda \mathcal{L}_{\text{pruning}}(\theta^*)
\end{equation}
where \( \mathcal{L}_{\text{task}} \) represents task-specific loss (e.g., classification), and \( \mathcal{L}_{\text{pruning}} \) regularizes sparsity. Common types of structured pruning include filter pruning~\cite{geometric,li2016pruning,C-SGD,AFP}, neuron pruning~\cite{lecun1989optimal,polar,yu2018nisp,unstructured_2019}, channel pruning~\cite{gao2021network,OBD-EigenD,Resrep,he2017channel}, and layer pruning~\cite{fan2019reducing,wang2018skipnet,unstructured_2017learning,elkerdawy2020filter}. These methods present unique challenges in balancing the trade-off between reduced size and maintaining task performance, with each approach requiring careful optimization to avoid excessive accuracy degradation.

\subsection{Torque-based Structured Pruning}
\begin{figure}[t] %H为当前位置，!htb为忽略美学标准，htbp为浮动图形
\centering %图片居中
\includegraphics[width=0.7\textwidth]{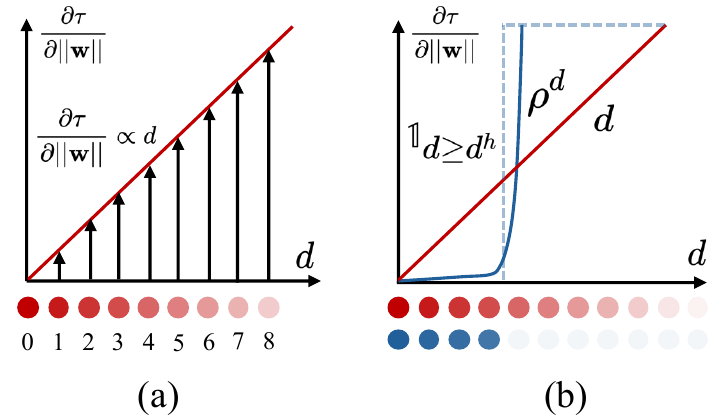} %插入图片，[]中设置图片大小，{}中是图片文件名
\caption{(a) Visualization of vanilla Torque-prune regularization. The circles below the coordinate system, arranged from left to right, represent neural modules at corresponding distances from the pivot point (leftmost circle). The depth of the circle's fill color represents the L2-norm of the module. The lighter the color, the lower the weight. (b) Comparison between the vanilla Torque-prune and ETP regularization.} %最终文档中希望显示的图片标题
% \vspace{-1.5em}
\label{fig:preli_method} %用于文内引用的标签
\end{figure}

Previous state-of-the-art pruning techniques require modifications to the network architecture or implementation of complex gradient
update rules. Whereas, Gupta \etal~\cite{efficientml5} propose a simple yet effective Torque-inspired approach (denoted as Torque in the following paper) which manages to achieve a great compression rate while requiring no change to model architecture and also very little or no fine-tuning. Concretely, analogous to the very definition of the physical concept (\ie~Torque), this approach proposes to apply a force to neural modules in order to consolidate the weights of a network layer around a selected pivot point during training. Formally, Torque approximates the concept with the following implementation:
\begin{equation}
    ||\mathbf{\tau}_i^l||_2=||\mathbf{F}_i^l \times \mathbf{r}_i^l||_2 \approx ||\mathbf{w}_i^l||_2 \cdot d_i^l,\ i\in \mathbb{Z}^+
\end{equation}
where $\mathbf{\tau}_i$ denotes the torque applied to the $i^{th}$ neural module of a layer $l$, $\mathbf{r}_i$ is the corresponding position vector. Torque-pruning approximates the L2-norm of the Torque that applies to the neural module's weights as the multiplication of the L2-norm of the module's weight matrix (\ie~force) $||\mathbf{w}_i^l||_2$ and the Euclidean distance of their corresponding indices $d_i^l=||\rho_i^l-\rho_p^l||_2$, where $\rho_i^l$ is the index of the $i^{th}$ neural module, $\rho_p^l$ is the index of the pivot point. Gupta \etal~\cite{efficientml5} adopt a random indexing strategy for the modules within a layer, which performs well empirically. Given the Torque $\mathbf{\tau}_i$, Torque-pruning proposes using it as a force that pushes the weights of neural modules that are distant from the pivot point to zero. Concretely, it implements it as a regularization term $\mathcal{L}_{\text{Torque}}$. The detailed optimization objective is as follows:
\begin{align}
    \mathcal{L}_{\text{total}}(\theta^*) & = \mathcal{L}_{\text{task}}(\mathcal{N}(\mathbf{x}; \theta^*)) + \lambda \mathcal{L}_{\text{Torque}}(\theta^*)\\
    &= \mathcal{L}_{\text{task}}(\mathcal{N}(\mathbf{x}; \theta^*)) + \lambda \sum_{l}\sum_{i}||\mathbf{\tau}_i^l||_2
\end{align} 
where $\lambda$ denotes the regularization coefficient of $\mathcal{L}_{\text{Torque}}$. \Cref{fig:preli_method}(a) shows an intuitive visualization of the vanilla Torque regularization. Specifically, the $\mathcal{L}_{\text{Torque}}$ regularization imposes a penalty on neural modules proportional to their distance from the pivot point (\ie~$\frac{\partial \tau}{\partial ||\mathbf{w}||}\propto ||\rho_i^l-\rho_p^l||_2$); the further it is, the more its weights would be penalized (illustrated via the depth of the color of the representing circles). 

\section{Exponential Torque Pruning}
In this section, we introduce the detailed motivation and design of our proposed method, called ETP (\underline{E}xponential \underline{T}orque \underline{P}runing). The key motivation of ETP is the sub-optimal force application scheme of the previous state-of-the-art Torque-prune approach. Specifically, we argue that the linear proportionality between the partial derivative of torque regularization with respect to the neural module's weights and the distance (\ie~$\frac{\partial \tau}{\partial ||\mathbf{w}||}\propto ||\rho_i^l-\rho_p^l||_2$) is inappropriate. Analogously, this denotes that for neural modules of different distances from the pivot point, we are applying the same amount of force to drive them to zero. As demonstrated in \Cref{fig:moti}, such linear force application scheme fails to constrain the weights of modules that are distant from the pivot point, while inappropriately penalizing the ones that are close and necessary for inference. 
To achieve a sparser and effective network architecture, we propose using a nonlinear force application scheme. Intuitively, we should apply a much larger force on the distant neural modules to drive them towards zero, while a smaller or no force (\ie~penalty) on those that are close to the pivot point, since they are essential for effective inference.
\Cref{fig:preli_method}(b) shows the intuitive visualization of different force application schemes as well as the illustration of the regularized neural modules' L2-norm of the corresponding schemes (denoted as circles filled with colors). 

Ideally, unlike the post-regularized neural modules of the vanilla Torque pruning, which are still dense in weights (illustrated by the red circles sequence), even though they are distant from the pivot point, we would like to achieve a sparse post-regularized network architecture (denoted as the blue circles sequence). Therefore, we formulate the nonlinear force application scheme using a Heaviside step function (denoted as the dashed blue line), such that $\frac{\partial \tau}{\partial ||\mathbf{w}||}\propto \mathbbm{1}_{||\rho_i^l-\rho_p^l||_2\geq d^h}$, where $d_h$ denotes the threshold distance from the pivot point, $\mathbbm{1}_{||\rho_i^l-\rho_p^l||_2\geq d^h}$ denotes the indicator function that outputs $1$ if the relative distance of the investigated neural module and the pivot point $||\rho_i^l-\rho_p^l||_2$ is larger than $d_h$, otherwise it outputs $0$. Thus, the detailed implementation of the Heaviside Torque regularization $||\tilde{\mathbf{\tau}}_i^l||_2$ is as follows:
\begin{align}
    ||\tilde{\mathbf{\tau}}_i^l||_2 & = ||\mathbf{F}_i^l \times \mathbf{r}_i^l||_2 \\ &\approx ||\mathbf{w}_i^l||_2 \cdot( \epsilon\cdot\mathbbm{1}_{||\rho_i^l-\rho_p^l||_2\geq d^h}),\ i\in \mathbb{Z}^+
\end{align}
Specifically, for neural modules within the threshold distance $d_h$, we apply zero force on them since they are considered necessary for inference, and for the modules that are beyond $d_h$, we exert a large force which is of size $\epsilon$ on these modules, driving their weights toward zero. Whereas the Heaviside step function is non-differentiable, we therefore use an exponential function for approximation (denoted as the solid blue line in \Cref{fig:preli_method}(b)), formally:
\begin{align}
    ||\hat{\mathbf{\tau}}_i^l||_2 & = ||\mathbf{F}_i^l \times \mathbf{r}_i^l||_2 \\ 
&\approx ||\mathbf{w}_i^l||_2 \cdot \lambda^{||\rho_i^l-\rho_p^l||_2},\ i\in\mathbb{Z}^+
\end{align}
where $\lambda$ is a hyperparameter that serves as the base of the exponentiation that controls the threshold distance. Finally, given the exponential approximation of the Heaviside Torque, the overall optimization objective with the exponential torque pruning regularization (ETP) loss is:
\begin{align}
    \mathcal{\hat{L}}_{\text{total}}(\mathbf{w}) & = \mathcal{L}_{\text{task}}(\mathbf{x}; \mathbf{w}) + \beta\cdot  \mathcal{L}_{\text{ETP}}^\mathbf{w}\\
    &= \mathcal{L}_{\text{task}}(\mathbf{x}; \mathbf{w}) + \beta \sum_{l}\sum_{i} ||\mathbf{w}_i^l||_2 \cdot  \lambda^{||\rho_i^l-\rho_p^l||_2}
\end{align} 
where $\mathcal{L}_{\text{task}}$ is the original optimization objective of the specific task, $\beta$ is the regulatory coefficient of the ETP loss.

\section{Experiments}
To evaluate the effectiveness of ETP, we conduct experiments on four distinct domains, including 
vision, language, graph, and time series. In the remainder of this section, we first introduce the detailed experimental setup, and then we 
answer three research questions (RQs) to lead our discussion, which are as follows: 

\textbf{(RQ1) Speed-up Improvement} How effective is ETP in improving the models' speed-up while retaining a low performance drop? 

\textbf{(RQ2) Aggresive Pruning Analysis} Does ETP perform well under different speed-up ratios?

\textbf{(RQ3) Effectiveness on Large Models} Can ETP effectively compress the prevalent large models?

\subsection{Experimental Setup}
We demonstrate the effectiveness of ETP by evaluating it on multiple benchmarks of different domains. The details of the benchmarks and the corresponding backbone models we evaluated on are as follows\footnote{Please refer to the appendix for the detailed experimental setup.}: 
% \vspace{-4mm}
\paragraph{Datasets \& Backbone Models.} 
For the image classification tasks, we evaluate on CIFAR-10, CIFAR-100~\cite{CIFAR}, and ImageNet~\cite{ImageNet} datasets. Both CIFAR-10 and CIFAR-100 consist of 50{,}000 training and 10{,}000 test images of size $32 \times 32$, with 10 and 100 classes respectively. All images are normalized using dataset-specific RGB means and standard deviations and resized to $224 \times 224$. For ImageNet, we use the ILSVRC-2012 subset with 1.2 million training images and 50{,}000 validation images across 1{,}000 classes. The images are resized and randomly cropped to $224 \times 224$ during training.  In terms of the backbone models, we follow the setup of Gupta~\etal\cite{efficientml5} and conduct experiments on the CNN models with linear connections only (\ie~VGG-19~\cite{VGG}), and the ones with residual connections (ResNet-50 and ResNet-56~\cite{ResNet}).
For the graph classification task, we use the Protein-Protein Interaction (PPI) dataset~\cite{ppi}, which contains 24 graphs with over 56,000 nodes and 818,000 edges, each node is represented with 50-dimensional features. We adopt the Graph Attention Network (GAT)~\cite{GAT} as the backbone model.
For the domain of natural language understanding (NLU), we evaluated two GLUE benchmark datasets: SST-2 and MRPC~\cite{glue}. SST-2 is a sentiment classification task with 67{,}349 training and 872 test examples, while MRPC is a paraphrase detection task with 3{,}668 training and 408 test pairs. We use BERT~\cite{devlin2019bert} and RoBERTa~\cite{liu2019roberta} models as the backbone models for evaluation.
For the time-series forecasting task, we use the ETTh1 dataset from the ETT benchmark suite~\cite{zhou2021informer}. The dataset includes hourly energy consumption features across one week, with 7 input features and 1 target variable. We adopt the Informer model~\cite{zhou2021informer} as the backbone, using an input sequence length of 96 and forecasting 48 future time steps. The features are normalized using the z-score normalization based on the training data statistics.
% PUBMED and CORA for graph classification, and SST-2, MRPC, and RTE for language classification. Additionally, we extend our study to time-series forecasting using the ETTH1 dataset.  

% For model evaluation, we employ several widely-used architectures, including VGG and ResNet for vision tasks, Graph Attention Networks for graph-based tasks, and BERT for language classification. For time-series forecasting, we utilize Informer. The language classification experiments leverage pre-trained models from Hugging Face, while for other tasks, model training and pruning are performed simultaneously. Following pruning, all models undergo a standardized fine-tuning process to ensure fair comparison.  
% \vspace{-4mm}
\paragraph{Compared Methods.}
We mainly compare ETP with the vanilla Torque pruning~\cite{torque} and DepGraph~\cite{fang2023depgraph}, a general-purpose structural pruning framework that constructs a dependency graph to automatically group and prune structurally coupled parameters across diverse architectures such as CNNs, Transformers, and GNNs. Apart from these methods, on the vision datasets, we also compare ETP against HRank~\cite{hrank}, SFP~\cite{SFP}, and GReg~\cite{GReg}. HRank prunes filters with low-rank feature maps based on the observation that rank correlates with information richness. SFP allows pruned filters to be updated during training, preserving model capacity; whereas GReg employs a growing L2 regularization scheme to simultaneously improve pruning schedules and exploit Hessian-informed importance scoring. %For CIFAR-10 (ResNet-56), we additionally benchmark against DepGraph w/o SL, a variant of DepGraph~\cite{fang2023depgraph} introduced in the original paper. In this setting, no sparsity-inducing regularization is applied during training. Instead, pruning is performed based on an importance metric computed using the dependency graph, followed by standard fine-tuning.

% \vspace{-4mm}
\paragraph{Evaluation Measurements.}
An ideal model compression algorithm should control the compression-accuracy tradeoff well, \ie~that is to effectively reduce the models' size, therefore reducing the number of computations for inference while controlling the accuracy loss within an acceptable range. To evaluate the compression-accuracy tradeoff quantitatively, we follow previous literature~\cite{fang2023depgraph,torque,GReg} and measure the \texttt{speed-up} and \texttt{accuracy-drop} metrics simultaneously. Specifically, speedup is defined as follows:  $\texttt{speed-up} = \frac{\texttt{MACs}_{\text{base}}}{\texttt{MACs}_{\text{pruned}}}$, MACS (Multiply-Accumulate operations) denotes the total number of arithmetic operations required for a single forward pass of the model. This is often used to
\begin{wraptable}{t}{0.6\textwidth}
% \vspace{-4mm}
\centering
\caption{Pruning results on vision benchmarks. We highlight the top-1 and top-2 results in red and blue respectively.}
\label{tab:vision}
\resizebox{0.6\textwidth}{!}{
\begin{tabular}{lllll} 
\toprule
\multicolumn{5}{c}{\textbf{CIFAR10 (ResNet56)}}\\ 
\hline\hline
\textbf{Method} & \textbf{Base} & \textbf{Pruned}  &\textbf{Acc. Drop} & \textbf{Speed-up} \\
\hline
% HRank~           & 93.26 & 93.17  &-0.09 & 2.00$\times$ \\
% SFP~             & 93.59 & 93.36  &-0.23 & 2.11$\times$ \\
% Torque (p)~      & 93.48 & \textcolor{red}{\textbf{93.76}}  &\textcolor{red}{\textbf{+0.28}} & 2.15$\times$ \\
% DepGraph~        & 93.53 & \textcolor{blue}{\textbf{93.77}}&\textcolor{blue}{\textbf{+0.24}} & 2.11$\times$ \\
% \rowcolor{mygray}
% \textbf{ETP (Ours)}      & 93.44 & 93.56  &+0.12 & 2.20$\times$ \\ 
% \hdashline
GReg-1~          & 93.36 & 93.18  &-0.18 & 2.55$\times$ \\
GReg-2~          & 93.36 & 93.36  &+0.00 & 2.55$\times$ \\
DepGraph~        & 93.53 & \textcolor{blue}{\textbf{93.64}}  &\textcolor{blue}{\textbf{+0.11}} & 2.57$\times$ \\
 Torque (p)& 93.48& 93.26& -0.22&2.72$\times$\\
\rowcolor{mygray}
\textbf{ETP (Ours)} & 93.44 & \textcolor{red}{\textbf{93.66}}  &\textcolor{red}{\textbf{+0.22}} & 2.93$\times$ \\
\hline
\multicolumn{5}{c}{\textbf{ImageNet (ResNet50)}}\\
\hline\hline
\textbf{Method} & \textbf{Base} & \textbf{Pruned}  &\textbf{Acc. Drop} & \textbf{Speed-up} \\
\hline
HRank~     & 76.15 & 74.98  &-1.17 & 1.78$\times$ \\
SFP~       & 76.15 & 74.51  &-1.64 & 1.72$\times$ \\
GReg-2~    & 76.13 & 75.36  &-0.77 & 1.49$\times$ \\
Depgraph~  & 76.15 & \textcolor{red}{\textbf{75.83}}  &\textcolor{red}{\textbf{-0.32}} & 2.08$\times$ \\
Torque (p)~& 76.07 & 74.67  &-1.40 & 2.34$\times$ \\
\rowcolor{mygray}
\textbf{ETP (Ours)} & 76.15 & \textcolor{blue}{\textbf{75.62}}  &\textcolor{blue}{\textbf{-0.53}} & 2.30$\times$ \\
\hline
\multicolumn{5}{c}{\textbf{CIFAR100 (VGG19)}}\\
\hline\hline
\textbf{Method} & \textbf{Base} & \textbf{Pruned}  &\textbf{Acc. Drop} & \textbf{Speed-up} \\
\hline
GReg-1~          & 74.02 & 67.35  &-6.67 & 8.84$\times$ \\
GReg-2~          & 74.02 & 67.75  &-6.27 & 8.84$\times$ \\
Depgraph~        & 73.50 & \textcolor{blue}{\textbf{70.39}}  &\textcolor{blue}{\textbf{-3.11}} & 8.92$\times$ \\
Torque (r)~      & 73.03 & 65.87  &-7.16 & 8.88$\times$ \\
\rowcolor{mygray}
\textbf{ETP (Ours)} & 73.50 & \textcolor{red}{\textbf{71.30}}  &\textcolor{red}{\textbf{-2.20}} & 9.03$\times$ \\
\bottomrule
\end{tabular}
}
% \vspace{-4mm}
\end{wraptable}
approximate computational cost and inference latency. Intuitively, \texttt{speed-up} quantifies how much more efficient the pruned model is compared to the original model. A higher value indicates greater computational savings, which enables faster inference and lower energy consumption.
The accuracy drop is defined as: 
$\texttt{accuracy-drop} = \texttt{accuracy}_{\text{pruned}} - \texttt{accuracy}_{\text{base}}$
Concretely, the $\texttt{accuracy-drop}$ measures the accuracy loss of the model before and after pruning. Other task specific metrics are further elaborated in Appendix. %\cref{subsec:task-specific-metrics}.

% In addition to speed-up and accuracy drop, we employ the following task-specific evaluation metrics tailored to each domain:
% \textbf{F1 Score (GAT on PPI):} For multi-label node classification tasks, we report the F1 score, defined as the harmonic mean of precision and recall:
%     \[
%         \text{F1} = 2 \cdot \frac{\text{Precision} \cdot \text{Recall}}{\text{Precision} + \text{Recall}}.
%     \]
% \textbf{MAE and MSE (Informer on ETTh1(48)):} For time-series forecasting, we evaluate performance using:
%     \begin{align*}
%         \text{MAE} &= \frac{1}{n} \sum_{i=1}^{n} |y_i - \hat{y}_i|, \\
%         \text{MSE} &= \frac{1}{n} \sum_{i=1}^{n} (y_i - \hat{y}_i)^2,
%     \end{align*}
%     where $y_i$ and $\hat{y}_i$ denote the ground truth and predicted values, respectively.\\
% \textbf{BLEU Score (VLM on Flickr8k):} For image captioning, we report BLEU-$n$ scores ($n=1$ to $4$), which compute the geometric mean of $n$-gram precisions with a brevity penalty:
%     \[
%         \text{BLEU} = \text{BP} \cdot \exp\left( \sum_{n=1}^{N} w_n \log p_n \right),
%     \]
%     where $p_n$ denotes $n$-gram precision, $w_n$ are uniform weights, and $\text{BP}$ is the brevity penalty to penalize short hypotheses. BLEU is reported as a percentage for readability.
\begin{figure}[t] %H为当前位置，!htb为忽略美学标准，htbp为浮动图形

\centering %图片居中
\includegraphics[width=0.99\textwidth]{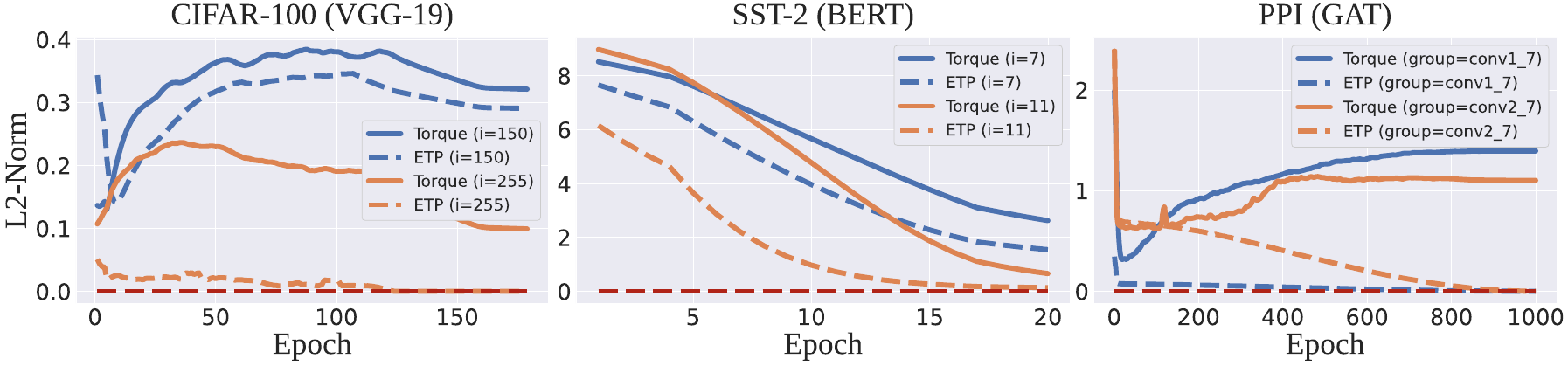} %插入图片，[]中设置图片大小，{}中是图片文件名
\caption{Comparison of the L2-norm curves during the training process.} %最终文档中希望显示的图片标题
% \vspace{-1.5em}
\label{fig:l2} %用于文内引用的标签
\end{figure}
\vspace{-4mm}
\subsection{Speed-up improvement (RQ1)}
To answer RQ1, we compare ETP against the state-of-the-art pruning techniques on four domains (\ie~image classification, graph classification, natural language understanding (NLU), and time-series forecasting). The detailed results are illustrated in \Cref{tab:vision}, \Cref{tab:nlu}, \Cref{tab:Informer}, and \Cref{tab:GAT}. 
First, for image classification tasks, we compare ETP with the state-of-the-art pruning baselines on CIFAR-10 with ResNet-56 model, ImageNet on ResNet-50, and CIFAR-100 with VGG-19 model, following the same experimental setup as previous literature~\cite{fang2023depgraph,torque}. The results in \Cref{tab:vision} show that ETP manages to achieve comparative or slightly better performance than that of the previous state-of-the-art on ResNet-56 and ResNet-50 models, 
while it significantly surpasses the baselines on VGG-19. Specifically, on the CIFAR-100 dataset using the VGG-19 backbone model, ETP achieves a $9\times$ computational speed-up while incurring only a 2.2\% drop in classification accuracy, while Torque and DepGraph suffers from a significantly higher accuracy degradation of 7.16\% and 3.11\% respectively. 
We believe the more significant superiority of ETP on larger models (\eg~VGG19) over smaller models (\eg~ResNet) is because these models contain a higher degree of parameter redundancy, which leave space for a higher compression rate (\ie~speed-up) while retaining low accuracy drop, on which a more effective force application scheme can better demonstrate its validity. 
% We believe this is because for the low speed-up settings, the amount of parameters that require pruning is small, our exponential force application scheme is theoretically close to 
% We hypothesize that this is because VGG-19 contains a higher degree of parameter redundancy. This is due to its deep feedforward architecture and uniform layer structure which are usually easier to prune~\cite{han2015deep}. In contrast, ResNet architectures intrinsically constraint pruning due to multi-branch structure. The skip connections create inter-layer dependency that complicates structured pruning.~\cite{ResNet} Hence due to this structure that restricts pruning, ResNet becomes a very tricky model to prune and high speed-ups are not practically attainable by structured pruning\lzm{not very convincing}. 
% Consequently, ETP shows only comparable or slightly improved performance on these architectures as there is not much room for compression. 
% However, its consistent performance across model families highlights its robustness and general applicability.  

For the graph classification, NLU, and time-series forecasting tasks, we observe that the ETP majorly surpasses the previous state-of-the-art baselines by a significant margin. Specifically, for the GAT model on the PPI dataset, we evaluate different methods under two speed-up settings. For speed-up=$12\times$, ETP achieves a F1 score drop of only 0.027, while DepGraph incurs a F1 score drop of 0.03 for the same speed-up. The results on $9\times$ speed-up rate are similar. Note that though we have tried our best, the vanilla Torque method fails to achieve the 12$\times$ speed-up on the GAT (PPI). We believe that this is because Torque is unable to penalize the GAT models' parameters enough to make them structurally sparse. 
% The GAT architecture has 8 attention heads in the first 2 layers which are penalized as grouped parameters. The distance coefficient of Torque method scales to only 8 times from the 1st to the last attention head which is not enough to achieve the desired sparsity even after significantly increasing the weight of the torque loss compared to task loss. 
\begin{wraptable}{t!}{0.59\textwidth}
\caption{Pruning results on the Informer model.}
\label{tab:Informer}
\arrayrulecolor{black}
\resizebox{0.59\textwidth}{!}{
\begin{tabular}{ccccccc}
\hline
\multicolumn{7}{c}{\textbf{Etth-1(48) (Informer)}}                                                                                                                                                      \\ \hline \hline
\multicolumn{1}{c|}{}             & \multicolumn{2}{c|}{DepGraph}                              & \multicolumn{2}{c|}{Torque}          & \multicolumn{2}{c}{\cellcolor{mygray}\textbf{ETP (Ours)}} \\ \hline
\multicolumn{1}{c|}{Speed-Up}     & MAE           & \multicolumn{1}{c|}{MSE          }    & MAE    & \multicolumn{1}{c|}{MSE}    & \cellcolor{mygray}MAE    & \cellcolor{mygray}MSE\\ \hline
\multicolumn{1}{c|}{1$\times$}  & 0.319        & \multicolumn{1}{c|}{0.158       } & 0.319 & \multicolumn{1}{c|} {0.158} & \cellcolor{mygray}0.319 & \cellcolor{mygray}0.158\\
\multicolumn{1}{c|}{2.5$\times$}  & 0.3559        & \multicolumn{1}{c|}{0.1636       } & \textcolor{red}{\textbf{0.3398}} & \multicolumn{1}{c|}{0.1621} & \cellcolor{mygray}0.3402 & \cellcolor{mygray}\textcolor{red}{\textbf{0.1618}}\\
\multicolumn{1}{c|}{4$\times$}    & 0.3632        & \multicolumn{1}{c|}{0.1671       } & \textcolor{red}{\textbf{0.3492}} & \multicolumn{1}{c|}{0.1665} & \cellcolor{mygray}0.3495 & \cellcolor{mygray}\textcolor{red}{\textbf{0.1631}}\\
\multicolumn{1}{c|}{6.5$\times$}  & 0.3737        & \multicolumn{1}{c|}{0.1702       } & 0.3606 & \multicolumn{1}{c|}{0.1698} & \cellcolor{mygray}\textcolor{red}{\textbf{0.3580}} & \cellcolor{mygray}\textcolor{red}{\textbf{0.1645}}\\
\multicolumn{1}{c|}{10.5$\times$} & 0.3818        & \multicolumn{1}{c|}{0.1743       } & 0.3723 & \multicolumn{1}{c|}{0.1756} & \cellcolor{mygray}\textcolor{red}{\textbf{0.3643}} & \cellcolor{mygray}\textcolor{red}{\textbf{0.1661}}\\
\multicolumn{1}{c|}{14.5$\times$} & 0.3959        & \multicolumn{1}{c|}{0.1797       } & 0.3843 & \multicolumn{1}{c|}{0.1810} & \cellcolor{mygray}\textcolor{red}{\textbf{0.3726}} & \cellcolor{mygray}\textcolor{red}{\textbf{0.1678}}\\
\multicolumn{1}{c|}{25$\times$}   & 0.4118        & \multicolumn{1}{c|}{0.1852       } & 0.3937 & \multicolumn{1}{c|}{0.1843} & \cellcolor{mygray}\textcolor{red}{\textbf{0.3812}} & \cellcolor{mygray}\textcolor{red}{\textbf{0.1692}}\\ \hline
\end{tabular}
% \vspace{-4mm}
}
\arrayrulecolor{black}
\end{wraptable}
ETP consistently outperforms DepGraph and Torque on Informer for Etth-1 dataset as well for $\texttt{speed-up} \ge 6.5 \times$ for both MAE and MSE. For $\texttt{speed-up} \le 4 \times$, ETP consistently outperforms DepGraph with a significant performance gain and is competent or slightly worse than Torque. 
The results indicate that ETP's superiority is more significant under large speed-up rate scenarios. The reason is that with
more redundant parameters (\ie~more distant and redundant neural modules) awaiting to be pruned, 
ETP can achieve more effective pruning by applying much larger penalty on the redundancy while retaining the modules that are necessary for inference according to its exponential force application scheme (we further validate this conclusion in RQ2 via the aggressive pruning analysis).
\begin{table}
\centering
\caption{Pruning results on the NLU benchmarks.}
\label{tab:nlu}
\resizebox{0.9\textwidth}{!}{
\begin{tabular}{ccccc|cccc} 
\toprule
\multicolumn{9}{c}{\textbf{SST-2}} \\ 
\hline\hline
\multirow{2}{*}{\textbf{Method}} & \multicolumn{4}{c|}{\textbf{BERT}}                                                                                     & \multicolumn{4}{c}{\textbf{RoBERTa}} \\ 
\cline{2-9}
                                 & \textbf{Base}          & \textbf{Pruned}          & \textbf{Acc. Drop}          & \textbf{Speed-up}          & \textbf{Base}          & \textbf{Pruned}          & \textbf{Acc. Drop}          & \textbf{Speed-up} \\ 
\hline
DepGraph                         & 93.5\%                 & 91.8\%                   & -1.7\%                      & 11$\times$                 & 95.3\%                 & 89.9\%                   & -5.4\%                      & 13.5$\times$ \\
Torque                           & 93.5\%                 & 90.9\%                   & -2.6\%                      & 11$\times$                 & 95.3\%                 & 90.6\%                   & -4.7\%                      & 13.5$\times$ \\
\rowcolor{mygray}
\textbf{ETP (Ours)}              & 93.5\%                 & \textcolor{red}{\textbf{92.1\%}}                   & \textcolor{red}{\textbf{-1.4\%}}                      & 11$\times$                 & 95.3\%                 & \textcolor{red}{\textbf{92.9\%}}                   & \textcolor{red}{\textbf{-2.4\%}}                      & 13.5$\times$ \\ 
\hline
\multicolumn{9}{c}{\textbf{MRPC}} \\ 
\hline\hline
\multirow{2}{*}{\textbf{Method}} & \multicolumn{4}{c|}{\textbf{BERT }}                                                                           & \multicolumn{4}{c}{\textbf{RoBERTa }} \\ 
\cline{2-9}
                                 & \textbf{Base}          & \textbf{Pruned}          & \textbf{Acc. Drop}          & \textbf{Speed-up}          & \textbf{Base}          & \textbf{Pruned}          & \textbf{Acc. Drop}          & \textbf{Speed-up} \\
\hline
DepGraph                         & 88.0\%                 & 83.5\%                   & -4.5\%                      & 8$\times$                  & 90.0\%                 & 86.1\%                   & -3.9\%                      & 8$\times$ \\
Torque                           & 88.0\%                 & 83.2\%                   & -4.8\%                      & 8$\times$                  & 90.0\%                 & 85.3\%                   & -4.7\%                      & 8$\times$ \\
\rowcolor{mygray}
\textbf{ETP (Ours)}              & 88.0\%                 & \textcolor{red}{\textbf{85.0\%}}                   & \textcolor{red}{\textbf{-3.0\%}}                      & 8$\times$                  & 90.0\%                 & \textcolor{red}{\textbf{86.6\%}}                   & \textcolor{red}{\textbf{-3.4\% }}                     & 8$\times$ \\
\bottomrule
\end{tabular}
}
\vspace{-4mm}
\end{table}

\begin{table}
\centering
\caption{Pruning results on the GAT model.}
\label{tab:GAT}
\arrayrulecolor{black}
\resizebox{0.58\textwidth}{!}{
\begin{tabular}{ccccc} 
\toprule
\multicolumn{5}{c}{\textbf{PPI (GAT)}}                                                                                               \\ 
\hline\hline
\textbf{Method}     & \textbf{Base} & \textbf{Pruned} & \textbf{F1 score Drop} & \textbf{Speed-Up}  \\ 
\hline
% DepGraph+Random   & 0.986  & 0.951 & -0.035  & 8.05$\times$   \\
% DepGraph+CP     & 0.986  & 0.957 & -0.029&  8.05$\times$  \\
% DepGraph w/o SL & 0.986  & 0.953 & -0.033&  8.26$\times$  \\
DepGraph & 0.986 & 0.961 & -0.025  & 8.43$\times$  \\
 Torque& 0.986& -& -&-\\
\rowcolor{mygray}
\textbf{ETP (Ours)}    & 0.986 & \textcolor{red}{\textbf{0.963}}  & \textcolor{red}{\textbf{-0.023}}  & 9.13$\times$\\ \hdashline
DepGraph & 0.986 & 0.956 & -0.030  & 12$\times$  \\
 Torque& 0.986& -& -&-\\
\rowcolor{mygray}
\textbf{ETP (Ours)}    & 0.986 & \textcolor{red}{\textbf{0.959}}  & \textcolor{red}{\textbf{-0.027}}  & 12.16$\times$\\
\bottomrule
\end{tabular}
}
\arrayrulecolor{black}
\end{table}
To better understand the source of improvement, 
we conduct an in-depth analysis to track the progress of the L2-norm of specific neural modules during the learning process. Concretely, we randomly select two neural modules within a specific layer that are of different distances from the pivot point, 
we compare the L2-norm learning process of ETP and the vanilla Torque pruning approach. The results are shown in \Cref{fig:l2}. We can observe that compared with the vanilla Torque, ETP can significantly reduce the L2-norm of the distant neural modules, \eg~for VGG-19 trained on CIFAR-100, ETP manages to optimally prune the distant module (\ie~L2-norm equals to 0 ($||m^l_{254}||=0.0$)), while Torque remains a high L2-norm (\ie~$||m^l_{254}||= 0.134$). The extensive L2-norm analysis during training validates that the exponential force application scheme enables ETP to achieve a significantly sparser yet effective neural network architecture, resulting in a much higher compression rate with minimal performance degradation.

% \begin{wrapfigure}{}{0.5\textwidth}
% \vspace{-4mm}
% \centering
% \caption{Pruning results on NLU benchmarks}
% \label{tab:language}
% \resizebox{0.5\textwidth}{!}{
% \begin{tabular}{cccccc}
% \hline
% \multicolumn{6}{c}{NLU Task}                                                                                                                               \\ \hline
% \multicolumn{1}{c|}{Dataset} & \multicolumn{1}{c|}{Model}   & \multicolumn{1}{c|}{Speed-Up}     & \multicolumn{3}{c}{Acc. Drop}                            \\ \cline{4-6} 
% \multicolumn{1}{c|}{}        & \multicolumn{1}{c|}{}        & \multicolumn{1}{c|}{}             & DepGraph & Torque   & \cellcolor{mygray}ETP (Ours) \\ \hline
% \multicolumn{1}{c|}{SST-2}   & \multicolumn{1}{c|}{BERT}    & \multicolumn{1}{c|}{11$\times$}   & -1.7$\%$ & -2.6$\%$ & \cellcolor{mygray}-1.4$\%$   \\
% \multicolumn{1}{c|}{}        & \multicolumn{1}{c|}{RoBERTa} & \multicolumn{1}{c|}{13.5$\times$} & -5.4$\%$ & -4.7$\%$ & \cellcolor{mygray}-2.4$\%$   \\ \hline
% \multicolumn{1}{c|}{MRPC}    & \multicolumn{1}{c|}{BERT}    & \multicolumn{1}{c|}{8$\times$}    & -4.5$\%$ & -4.8$\%$ & \cellcolor{mygray}-3$\%$     \\
% \multicolumn{1}{c|}{}        & \multicolumn{1}{c|}{RoBERTa} & \multicolumn{1}{c|}{8$\times$}    & -3.9$\%$ & -4.7$\%$ & \cellcolor{mygray}-3.4$\%$   \\ \bottomrule
% \end{tabular}
% }
% %\vspace{-8mm}
% \end{wrapfigure}

\subsection{Aggressive Pruning Analysis (RQ2)}
\begin{figure}
    \centering
    \includegraphics[width=0.99\linewidth]{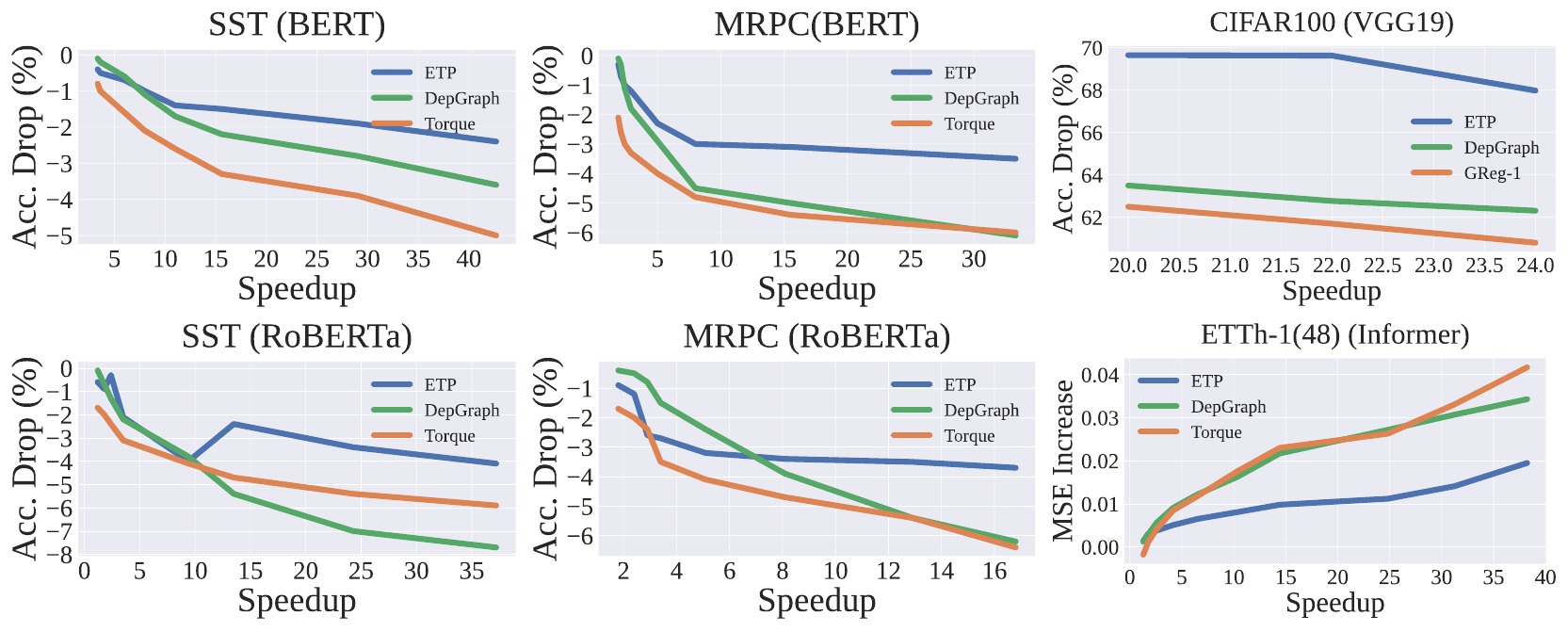}
    \caption{Results of aggressive pruning analysis for six distinct tasks.}
    \label{fig:NLU_agg}
\end{figure}

Different real-world applications require different levels of model compression due to hardware limitations; therefore, to perform well (\ie~retain low accuracy drop) under different speed-up ratios is a crucial ability for the model compression techniques. To systematically quantify such ability, we propose evaluating different pruning methods via the \textbf{aggressive pruning analysis}. Concretely, that is to record the model's accuracy drop across progressively increasing speed-up ratios.
% \begin{wrapfigure}{r}{0.48\textwidth}
%     \vspace{-8mm}
%     \centering
%     \includegraphics[width=0.45\textwidth]{img/agg_speed.png}
%     \setlength{\abovecaptionskip}{1mm}  % reduce space above caption
%     \caption{ETP, DepGraph~\cite{fang2023depgraph}, and GReg-1~\cite{GReg} under aggressive pruning scenarios.}
%     \label{fig:agg_prune}
%     \vspace{-2mm}
% \end{wrapfigure}
We conduct the analysis on the six different tasks, the results are shown in \Cref{fig:NLU_agg}. It is obvious that ETP manages to retain the accuracy within an acceptable range while the other compared methods suffer from a significant accuracy drop. For example, for BERT on MRPC, under the $30\times$ speed-up ratio, ETP achieves an accuracy drop of only 3.5$\%$, while DepGraph and Torque's accuracy drop by 6$\%$ and 6.1$\%$ respectively. Similarly, for VGG19 on CIFAR 100 dataset, under the 23 $\times$ speed-up setting, ETP incurs an accuracy drop of only 3.87$\%$, while DepGraph and GReg's accuracy is 10.73$\%$ and 13$\%$ respectively. We observe similar trends on Informer for ETTh-1 (48) dataset as well. ETP incurs a change in MSE of 0.02 for a 38$\times$ speed-up, while DepGraph incurs a change in MSE of 0.032 for the same speed-up. Torque performs the worst out of the 3 methods at 38$\times$ speed-up and incurs a change in MSE of 0.041. The results demonstrate that, thanks to a more reasonable force application scheme, ETP is a much more robust pruning technique and it is more suitable for scenarios that require a large speed-up ratio (\eg~model deployment on edge devices with limited computing power) compared to the previous state-of-the-art baselines.

\subsection{Effectiveness on Large Models (RQ3)}
\begin{wraptable}{!t}{0.57\textwidth}
% \vspace{-4mm}
\caption{Pruning results on the BLIP model.}
\label{tab:vlm}
\arrayrulecolor{black}
\resizebox{0.57\textwidth}{!}{
\begin{tabular}{lllll} 
\toprule
\multicolumn{5}{c}{\textbf{Flickr8k (BLIP)}}                                                                                               \\ 
\hline\hline
\textbf{Method}     & \textbf{BLEU-1} & \textbf{BLEU-2} & \textbf{BLEU-3} & \textbf{BLEU-4}  \\ 
\hline
Base   & 72.02$\%$  & 43.11$\%$  & 22.96$\%$  & 29.97$\%$   \\
L1     & 66.12$\%$  & 36.79$\%$  & 18.53$\%$  & 26.01$\%$   \\
Torque & 67.24$\%$  & 38.85$\%$  & 20.51$\%$  & 27.44$\%$   \\
\rowcolor{mygray}
\textbf{ETP (Ours)}    & \textcolor{red}{\textbf{69.78$\%$}}  & \textcolor{red}{\textbf{40.99$\%$}}  & \textcolor{red}{\textbf{21.79$\%$}}  & \textcolor{red}{\textbf{28.90$\%$}}   \\
\bottomrule
\end{tabular}
}
\arrayrulecolor{black}
\end{wraptable}

We further conduct preliminary experiments on large-scale models to demonstrate the universality and future potential of ETP. Specifically, we evaluate our method on the BLIP~\cite{blip} vision-language model using the Flickr8k~\cite{flickr8k} dataset. Following prior work~\cite{bleu,bleu1,bleu2,bleu3,bleu4}, we use BLEU-1 to BLEU-4 as the evaluation metrics. In this work, we freeze the weights of the image encoder and only perform structured pruning exclusively on the language decoder as a proof of concept. We enforce a uniform pruning budget: Each method removes 20\% of the total trainable parameters.
We compare ETP against L1-regularization-based pruning~\cite{tibshirani1996regression} and the vanilla Torque pruning approach. As shown in \Cref{tab:vlm}, ETP consistently outperforms both baselines in terms of all metrics. Notably, ETP achieves 69.78$\%$ on BLEU-1, outperforming Torque by 2.54$\%$ and L1 by 3.66$\%$. On the more stringent BLEU-4 metric, ETP yields only a 1.07$\%$ drop from the base, whereas Torque and L1 suffer drops of 2.53$\%$ and 3.96$\%$ respectively. These results highlight ETP's superior ability to retain captioning quality under aggressive pruning. Overall, this experiment suggests that ETP is effective even on large models. We leave the comprehensive evaluation of ETP on other vision-language and large language models to future work.
\vspace{-4mm}

\section{Related Works}
Model pruning is a widely studied compression technique that removes redundant parameters from neural networks to improve efficiency. Pruning approaches are generally categorized into unstructured and structured methods. In this section, we briefly review these two lines of research.

\paragraph{Unstructured pruning} aims to remove individual weights in a network, typically based on magnitude-based heuristics or importance scores~\cite{lecun1989optimal,unstructured,unstructured_2017learning,unstructured_2019}. One of the seminal works in this area was introduced by Han et al.~\cite{han2015deep}, who proposed an iterative pruning framework that eliminates weights with small magnitudes and then retrains the network to recover any lost accuracy. This approach was shown to significantly reduce model size while maintaining competitive performance. Subsequent work has explored various extensions and refinements, such as dynamic sparsity during training, sensitivity-aware pruning, and second-order information (e.g., Optimal Brain Damage~\cite{lecun1989optimal}).  Despite achieving high sparsity levels, unstructured pruning often results in irregular and non-dense parameter distributions, which pose challenges for practical acceleration. Without specialized hardware or software support, these irregular sparse patterns offer limited improvements in actual inference time or energy efficiency~\cite{he2023structured}. Recent advances have shifted attention toward \textit{one-shot} and \textit{data-aware} pruning methods, which aim to identify promising sparse subnetworks without the need for iterative prune-retrain cycles. Notably, SNIP~\cite{lee2019snip} proposes a saliency-based criterion that prunes weights at initialization by estimating their influence on the loss function using a first-order Taylor expansion. Building on this, GraSP~\cite{wang2020picking} introduces a gradient preservation metric that retains weights crucial for maintaining informative gradients during early training. These approaches reduce computational cost by avoiding full pretraining, though they may underperform compared to iterative methods at extreme sparsity. 
%More recently, SynFlow~\cite{tanaka2020synflow} demonstrates that unstructured pruning can be conducted entirely without data, by propagating synthetic inputs to estimate layer-wise importance while avoiding layer collapse. These methods represent a growing effort to formalize pruning as a differentiable and initialization-aware process, improving reproducibility and generality across architectures. Despite these improvements, unstructured pruning remains constrained by the need for custom sparse kernels or compiler support to translate parameter sparsity into runtime speedups, limiting its deployment in real-world systems.

\paragraph{Structured pruning} focuses on removing higher-level structures, such as entire channels, filters, or even layers. This yields a compact and dense model architecture that is more compatible with conventional hardware and software frameworks.~\cite{he2023structured,AFP,AMC,Resrep,GBN,hrank}
Early approaches, such as that by Li et al.~\cite{li2016pruning}, prune filters in convolution layers based on their $\ell_1$ norm, under the assumption that filters with smaller norms contribute less to the final output. He et al.~\cite{he2017channel} proposed channel pruning guided by evaluating the change in loss when specific channels are removed, allowing for a more data-driven pruning strategy. These methods are typically followed by fine-tuning to restore the performance of the pruned network~\cite{hrank,fang2023depgraph}.
Recent advances have cast structured pruning as a learning or optimization problem. 
For example, Liu et al.~\cite{liu2017learning} introduced a differentiable pruning framework that employs soft masks applied to channels, enabling end-to-end learning of pruning decisions during training. This approach allows gradients to flow through pruning masks, resulting in more informed and effective pruning strategies. Reinforcement learning-based methods~\cite{AMC} have also been proposed, where an agent learns to prune structures based on performance rewards, automatically balancing accuracy and efficiency trade-offs. Other lines of work employ regularization techniques to encourage sparsity during training (e.g., group Lasso), or meta-learning approaches to adapt pruning strategies across tasks~\cite{GReg,torque}.
Moreover, there has been increasing interest in combining pruning with other model compression techniques, such as quantization and knowledge distillation, to maximize efficiency gains. Some recent works also explore dynamic pruning strategies, where the network structure adapts at inference time based on input complexity.

\section{Conclusion}
In this work, motivated by the observation that the vanilla Torque-based pruning still fails to achieve satisfying model sparsity, we introduce a simple yet effective pruning method called Exponential Torque Pruning (ETP) based on an exponential force application scheme. Concretely, ETP imposes a larger force on distant neural modules so as to constrain their weights to zero while preserving those that are close to the pivot point that are indispensable for effective inference. Experiments across four diverse downstream domains and multiple model architectures (including modern vision-language models) demonstrate that ETP significantly outperforms prior state-of-the-art pruning methods, achieving a much higher compression rate while maintaining considerably lower accuracy degradation. Regarding the limitation, we observe that ETP's improvement in low-speed-up scenarios is less significant than in high-speed-up scenarios. Future progress could be made in the force application scheme to mitigate this limitation. Besides, we also plan to apply ETP to other emerging architectures (\eg~diffusion models, Mixture-of-Experts architectures, \etc) to further assess its generalizability.

% We introduced SORTS (Self-Organizing Regularization for Trailing Sparsity), a novel structured pruning method that integrates sparsity directly into training, eliminating the need for multi-stage pruning. By leveraging Exponential Parameter Loss ($\mathcal{L}_{EPL}$), SORTS enables models to self-organize their sparsity, achieving high compression rates with minimal accuracy loss. We observe that it outperforms the State-Of-The-Art especially in high compression rates. 

% Our experiments demonstrate that SORTS outperforms existing methods, especially in aggressive pruning scenarios, maintaining 99.19\% accuracy on MNIST (DarkNet19) at 24× speed-up and 93.66\% accuracy on CIFAR-10 (ResNet-56) at 2.93× speed-up. Additionally, SORTS proves highly effective on BERT, achieving 42.7× speed-up with minimal accuracy drop.  

% With strong theoretical backing and broad applicability across CNNs, Transformers, and GNNs, SORTS offers a scalable, model-agnostic, and efficient pruning solution, making it ideal for real-world, resource-constrained environments. Future work could explore adaptive regularization and quantization integration to push efficiency even further.  

% \input{tex/appendix}
\bibliographystyle{plainnat}
\bibliography{references}

\end{document}

% This document was modified from the file originally made available by
% Pat Langley and Andrea Danyluk for ICML-2K. This version was created
% by Iain Murray in 2018, and modified by Alexandre Bouchard in
% 2019 and 2021 and by Csaba Szepesvari, Gang Niu and Sivan Sabato in 2022.
% Modified again in 2023 and 2024 by Sivan Sabato and Jonathan Scarlett.
% Previous contributors include Dan Roy, Lise Getoor and Tobias
% Scheffer, which was slightly modified from the 2010 version by
% Thorsten Joachims & Johannes Fuernkranz, slightly modified from the
% 2009 version by Kiri Wagstaff and Sam Roweis's 2008 version, which is
% slightly modified from Prasad Tadepalli's 2007 version which is a
% lightly changed version of the previous year's version by Andrew
% Moore, which was in turn edited from those of Kristian Kersting and
% Codrina Lauth. Alex Smola contributed to the algorithmic style files.